\documentclass[sigconf]{acmart}
\usepackage{colortbl}
\AtBeginDocument{%
  }

\copyrightyear{2025}
\acmYear{2025}
\setcopyright{acmlicensed}
\acmConference[MM '25] {Proceedings of the 33rd ACM International Conference on Multimedia}{October 27--31, 2025}{Dublin, Ireland.}
\acmBooktitle{Proceedings of the 33rd ACM International Conference on Multimedia (MM '25), October 27--31, 2025, Dublin, Ireland}
\acmISBN{979-8-4007-2035-2/2025/10}
\acmDOI{10.1145/3746027.3755289}

\begin{document}

\title{GLDesigner: Leveraging Multi-Modal LLMs as Designer for Enhanced Aesthetic Text Glyph Layouts}

\author{Junwen He}
\email{junwen.he@mail.dlut.edu.cn}
\orcid{0009-0005-9314-4242}
\affiliation{%
  \institution{Dalian University of Technology}
  \city{Dalian}
  \country{China}
}
\author{Yifan Wang}
\authornote{Corresponding author}
\email{wyfan@dlut.edu.cn}
\affiliation{%
  \institution{Dalian University of Technology}
  \city{Dalian}
  \country{China}
}
\author{Lijun Wang}
\email{ljwang@dlut.edu.cn}
\affiliation{%
  \institution{Dalian University of Technology}
  \city{Dalian}
  \country{China}
}
\author{Huchuan Lu}
\email{lhchuan@dlut.edu.cn}
\affiliation{%
  \institution{Dalian University of Technology}
  \city{Dalian}
  \country{China}
}

\author{Chenyang Li}
\email{lee.lcy@alibaba-inc.com}
\affiliation{%
  \institution{Alibaba Group}
  \city{Shenzhen}
  \country{China}
}
\author{Hanyuan Chen}
\email{hanyuan.chy@alibaba-inc.com}
\affiliation{%
  \institution{Alibaba Group}
  \city{Hangzhou}
  \country{China}
}
\author{Jinpeng Lan}
\email{lanjinpeng.ljp@alibaba-inc.com}
\affiliation{%
  \institution{Alibaba Group}
  \city{Shenzhen}
  \country{China}
}
\author{Jun-Yan He}
\email{leyuan.hjy@alibaba-inc.com}
\affiliation{%
  \institution{Alibaba Group}
  \city{Shenzhen}
  \country{China}
}
\author{Bin Luo}
\authornote{Project Lead}
\authornotemark[1]
\email{luobin1986@gmail.com}
\affiliation{%
  \institution{Kingdee Group}
  \city{Shenzhen}
  \country{China}
}
\author{Yifeng Geng}
\email{cangyu.gyf@alibaba-inc.com}
\affiliation{%
  \institution{Alibaba Group}
  \city{Beijing}
  \country{China}
}

\renewcommand{\shortauthors}{Junwen He et al.}

\begin{abstract}
Text logo design heavily relies on the creativity and expertise of professional designers, in which arranging element layouts is one of the most important procedures.
However, this specific task has received limited attention, often overshadowed by broader layout generation tasks such as document or poster design. 
In this paper, we propose a Vision-Language Model (VLM)-based framework that generates content-aware text logo layouts by integrating multi-modal inputs with user-defined constraints, enabling more flexible and robust layout generation for real-world applications.
We introduce two model techniques that reduce the computational cost for processing multiple glyph images simultaneously, without compromising performance.
To support instruction tuning of our model, we construct two extensive text logo datasets that are five times larger than existing public datasets.
In addition to geometric annotations (\textit{e.g.}, text masks and character recognition), our datasets include detailed layout descriptions in natural language, enabling the model to reason more effectively in handling complex designs and custom user inputs. 
Experimental results demonstrate the effectiveness of our proposed framework and datasets, outperforming existing methods on various benchmarks that assess geometric aesthetics and human preferences. 
\end{abstract}




\maketitle

\section{Introduction}
\label{sec:intro}
Automatic layout design in creative AI is gaining increasing attention for its potential to enhance efficiency and assist designers in the creative process.
Traditional rule-based methods often result in plain and aesthetically unappealing layouts, heavily relying on professional designers to achieve visually pleasing outcomes—making the process time-consuming and labor-intensive.
In particular, text logo layout design poses unique challenges and remains underexplored compared to general graphic layout tasks. It requires careful consideration of font style, size, texture, and the semantic importance of keywords to ensure that the final design is both visually compelling and free of glyph collisions.

\begin{figure}[tbp]
    \centering
    \includegraphics[width=\columnwidth]{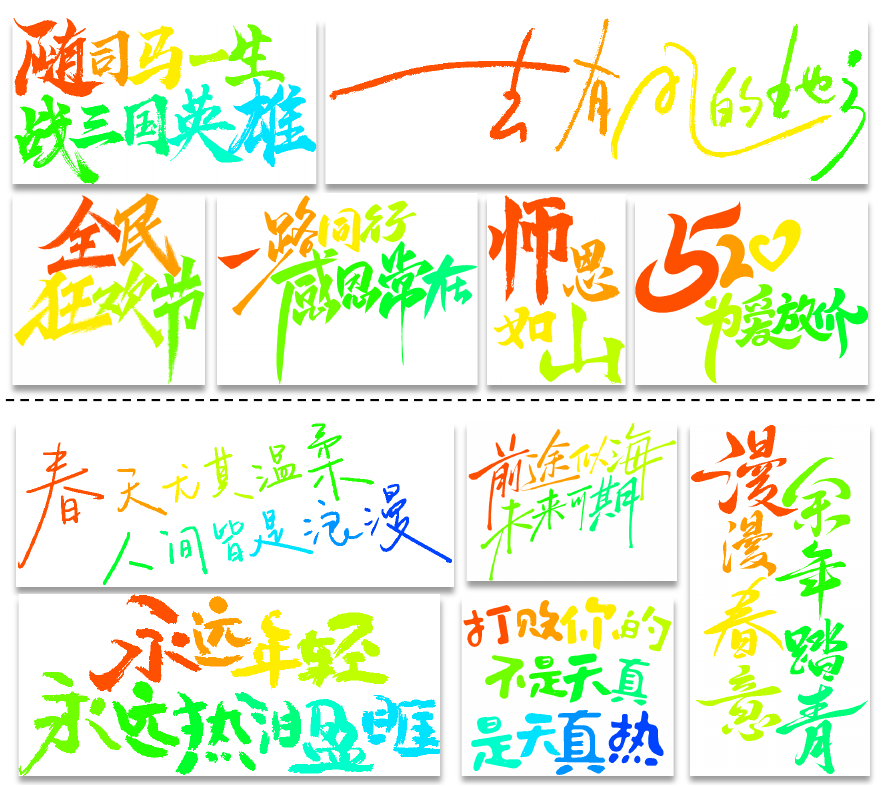}
    \caption{
    Illustrated are examples of text logos from our datasets, each glyph uniquely colored.
    We aim to enhance previous datasets by introducing more complex, compact layouts (top) and extended text sequences (bottom).
    }
    \label{intropdf}
\end{figure}

To overcome these challenges, recent approaches leverage generative models such as Generative Adversarial Networks (GANs) \cite{goodfellow2014gan} and Diffusion Models \cite{sohl2015diffusion} to synthesize graphic layouts by learning from examples of human-designed data.
The previous method \cite{wang2021textlogo} utilized GAN-based models to address this issue but remains limited in handling (1) long text sequences, (2) user-defined layout constraints, and (3) data variability and generality, making it less suitable for real-world design applications.
Additionally, the method encodes character semantics using pretrained semantic word embedding models \cite{li2018characterembed}, which are less expressive compared to contemporary large language models (LLMs) that are trained on trillions of text tokens.
Meanwhile, an increasing number of general layout generation methods \cite{feng2024layoutgpt, tang2023layoutnuwa, lin2023layoutprompter, yang2024posterllava} have begun exploring multi-modal, content-aware conditions ({\it{e.g.}}, background images, document contexts).
However, these approaches cannot be directly applied to the text logo layout task, as it requires managing multiple glyphs and incorporating more fine-grained details to ensure both aesthetic appeal and readability.

Acknowledging these limitations, we propose a unified VLM-based framework for text logo layout generation, leveraging the inherent capability of vision-language models to process multi-modal inputs, including glyph images and textual semantics.
Our approach builds upon pre-trained VLMs \cite{liu2023llava} with minimal architectural modifications—specifically, early-stage feature extraction to improve sensitivity to fine-grained textural details and mitigate glyph collisions.
To enhance computational efficiency, we apply adaptive average pooling, reducing the number of visual glyph tokens by a factor of $6^2$. This significantly accelerates both training and inference without introducing additional parameters or degrading performance.
The final layout outputs are represented as pure JSON texts, ensuring a standardized format that facilitates easy and stable post-rendering while enhancing training stability.
Furthermore, we fine-tune the language component using instruction-following techniques tailored to the text logo layout task, enabling the model to generate aesthetically pleasing designs by reasoning over glyph styles, semantic content, and optional user-defined constraints.

We construct two large-scale text logo datasets containing pixel-level annotations, detailed layout descriptions, and user-defined constraints, totaling over 17,000 samples — five times larger than the currently available public dataset \cite{wang2021textlogo}.
The datasets comprise synthesized logos generated using rule-based methods derived from expert designers, as well as real-world text logos collected from the internet, followed by systematic post-processing and annotation, as illustrated in Figure~\ref{intropdf}.
Extensive experiments demonstrate that leveraging both datasets significantly reduces glyph collisions and enhances layout creativity and diversity. 
Our proposed approach outperforms all existing methods across qualitative benchmarks and human preference studies, while also accommodating additional layout constraints—enabling more controllable and practical text logo generation in real-world scenarios.

The main contributions of this work are as follows:
\begin{enumerate}
    \item We propose a VLM-based framework for content-aware text logo layout generation, which integrates multi-modal inputs and optionally supports user-defined constraints via natural language instructions, enabling greater flexibility in real-world design scenarios.
    \item We construct two large-scale text logo datasets, comprising both synthesized and real-world samples, with fine-grained layout annotations and instruction-following data. These datasets significantly surpass existing ones in both scale and descriptive richness.
    \item Our method outperforms existing approaches across multiple benchmarks, effectively handling long text sequences and accommodating user-defined layout requirements. Furthermore, by incorporating glyph transformation and texture generation, our model demonstrates strong versatility in practical applications.
\end{enumerate}


\section{Related Work}
\label{sec:relatedwork}

\subsection{Automatic Layout Generation}

\noindent{\bf{Graphic Layout Generation}} encompasses neural networks to automatically arrange elements such as images, text, and backgrounds into visually appealing designs.
Early models like LayoutGAN \cite{li2019layoutgan} and LayoutVAE \cite{jyothi2019layoutvae} focused on document and poster layouts without considering specific content, relying on generative frameworks such as GANs \cite{goodfellow2014gan} and VAEs \cite{kingma2013vae}.
Recent advances using auto-regressive models \cite{arroyo2021variationaltransformer, gupta2021layouttransformer, kong2022blt} and diffusion-based approaches \cite{inoue2023layoutdm, zhang2023layoutdiffusion, hui2023unifyinglayoutdm} have significantly improved layout realism and diversity. 
However, these models often lack the content-awareness required for practical applications.
In response, content-aware layout generation has emerged to tackle more complex design scenarios, particularly in commercial poster design with non-empty backgrounds, often incorporating saliency maps for layout guidance \cite{hsu2023posterlayout, zhou2022cglgan}.
Conditional models further enhance layout prediction by integrating visual and textual encoders \cite{cao2022cvae, yu2022layoutdetr}, while other approaches utilize constrained optimization \cite{Kikuchi2021layoutgan++} and intermediate representations \cite{jiang2023layoutformer++, lin2023parseandplace} to better conform to user-defined constraints.

\noindent{\bf{Text Logo Layout Generation}} is a specialized domain within graphic design that focuses on the arrangement of pixel-level text glyphs.
This task demands meticulous attention to glyph stroke details and reading order to achieve both aesthetic harmony and legibility.
Currently, the only existing method \cite{wang2021textlogo} utilizes a GAN-based model to integrate both linguistic and visual information from input texts and glyph images to predict optimal layouts.
While this method shows promising results, it faces notable limitations in handling long text sequences, supporting user-defined constraints, and producing diverse layout variations—largely due to the scarcity of high-quality training data.

\begin{figure*}[ht]
    \centering
    \includegraphics[width=\textwidth]{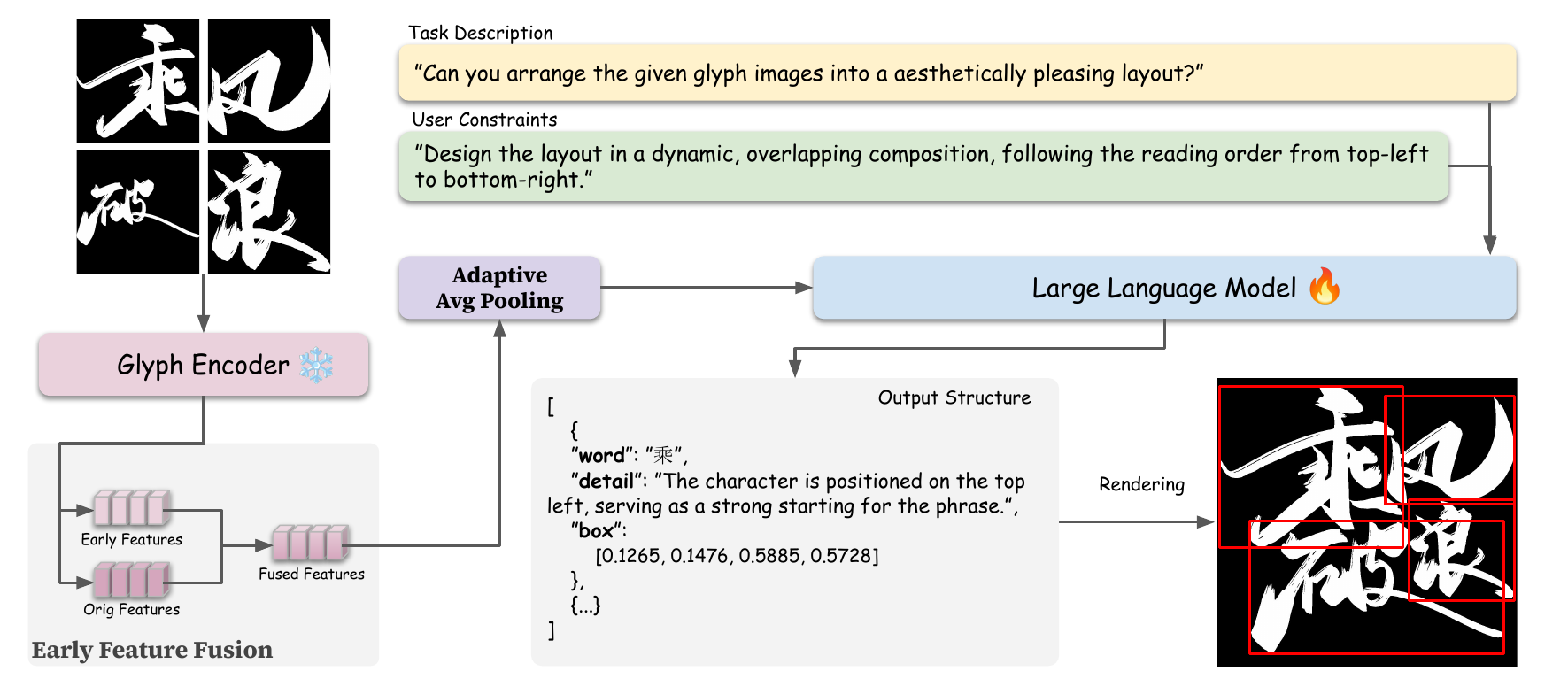}
    \caption{{\bf{Pipeline Overview of the Text Logo Layout Generation Method.}}
    For clarity and simplicity, vision-language projection layers and detailed prompt templates are omitted. 
    By incorporating {\bf Early Feature Fusion} and {\bf Adaptive Average Pooling}, as described in Section \ref{modelarchitecture}, the model ensures precise recognition of the glyph image’s textural details while reducing the number of visual tokens.
    The JSON-structured output guarantees a consistent and reliable rendering of the final text logo.
    }
    \label{network}
\end{figure*}

\subsection{Multi-modal Large Language Models}

Multi-modal Large Language Models, especially Vision-Language Models (VLMs), have excelled in numerous visual-language tasks.
By leveraging LLMs pretrained on extensive text corpora and using visual features as prompts through instruction tuning, these models achieve impressive performance in applications such as image captioning and Visual Question Answering (VQA).

While layouts, which can be encoded in coding formats such as HTML or JSON, are ideally suited to be processed by pre-trained Large Language Models (LLMs), as domain-specific data can be utilized to strength the code generation ability.
Following similar idea, several works use the same method to finetune VLLMs to have better performance on document layout understanding \cite{luo2024layoutllm, huang2022layoutlmv3}.
LayoutNUWA \cite{tang2023layoutnuwa} views layout generation as a code generation task and LayoutPrompter \cite{lin2023layoutprompter} leverages the training-free RAG pipeline to strengthen the in-context learning ability.
Recently, PosterLLaVA \cite{yang2024posterllava} employs structured text and visual instruction tuning to generate layouts under specific visual and textual constraints for poster designs.
Meanwhile, some contemporaneous works \cite{chen2024textdiffuser, chen2023textdiffuser2, feng2024layoutgpt} adopt LLMs as middle-ware for designing and direct layout-to-image generation through diffusion models.

However, these approaches cannot be directly applied to text logo layout generation, because they only consider a single background image and lack of consideration of fine-grained glyph stroke details.
We propose {\bf{GLDesigner}} to gap this bridge by combining all above advantages, and stimulate MLLM designing ability from fine-tuning on our large-scale constructed datasets.

\section{Method}
\subsection{Preliminaries}
\label{preliminaries}

A text logo can be represented as a collection of $N$ glyph elements, denoted by $\{(g_i, b_i)\}^{N}_{i=0}$.
Each $g_i \in \mathbb{R}^{H_i \times W_i}$ represents the raw visual component of the $i$-th glyph in image form, where $H_i$ and $W_i$ denote its height and width, respectively.
The corresponding bounding box $b_i = (x_i, y_i, x_i', y_i')$ defines the glyph’s position, with $(x_i, y_i)$ and $(x_i', y_i')$ representing the top-left and bottom-right coordinates of the box.
These coordinates are normalized with respect to the overall width and height of the canvas, ensuring resolution-independent and consistent placement across varying background sizes.

In the task of generating text logo layouts, the inputs consist of each glyph's character text $T_i$, its corresponding glyph image $g_i$, and optionally, user-defined constraints $c$.
The goal is to predict the positional bounding box $b_i$ for each glyph, based on these inputs, such that the resulting layout is visually coherent, semantically meaningful, and adheres to any specified constraints.

\subsection{Model Architecture}
\label{modelarchitecture}
We present an overview of our layout generation pipeline in Figure~\ref{network}, with detailed explanations provided in the following subsections.
Our model architecture is built upon LLaVA \cite{liu2023llava}, which enables unified processing of glyph images and textual inputs.
Specifically, we employ ViT-L \cite{radford2021clip} as the glyph encoder and LLaMA \cite{touvron2023llama2} as the language model.
To further enhance the model’s performance, we introduce two key techniques: {\bf Early Feature Fusion} and {\bf Adaptive Average Pooling}.
Early Feature Fusion improves the model’s ability to capture and interpret fine-grained glyph details, including stroke patterns and visual characteristics.
Adaptive Average Pooling, on the other hand, reduces the number of visual tokens by downsampling glyph image representations, significantly improving computational efficiency when handling multiple glyphs.
Together, these enhancements enable our model to more effectively understand visual-textual inputs and generate precise, aesthetically pleasing layout configurations.

\subsubsection{{\bf Early Feature Fusion.}}
\label{eff}
The default configuration of LLaVA \cite{liu2023llava} utilizes visual features from CLIP's \cite{radford2021clip} second-to-last layer for vision-language feature projection.
While these features capture the global properties and semantics of glyph images effectively, they tend to overlook finer textural details present in earlier layers.
To address this, we propose fusing early visual features into the original feature set. Based on experimental results, we select the second-to-last layer of CLIP as the point of fusion.
Additionally, we introduce a trainable two-layer multi-layer perceptron (MLP) as a projection layer, initialized to zero to ensure training stability.

\subsubsection{{\bf Adaptive Average Pooling.}}
\label{aap}
Another limitation of the original model is the linear increase in the number of visual tokens when processing multiple glyph images simultaneously.
As the number of glyph images grows, the total number of tokens can quickly exceed the model’s maximum sequence length, which in turn leads to a reduction in both training and inference efficiency.

To address this issue, we employ adaptive average pooling to reduce the number of visual tokens, without introducing any additional parameters beyond the existing trainable modules \cite{li2023blip2, Qwen-VL}.
As demonstrated in \cite{xu2024pllava, yao2024deco}, adaptive average pooling is a parameter-free technique that effectively preserves spatial locality while retaining key visual features.
Our experimental results further confirm that this approach significantly reduces computational overhead during both the training and inference phases, all while maintaining the quality of layout generation.

\subsection{Layout Structure}

We introduce customized prompt templates and a structured textual layout representation in JSON format, specifically designed for the instruction-following fine-tuning stage.
The details of the template are provided in Table~\ref{layoutstructure}.

To effectively guide the model, we begin with a detailed task description in the instruction prompt, which serves to clearly indicate and specify both the task content and the desired output format.
For handling glyph images alongside their textual semantics, each glyph is represented by unified tokens as $T_i$: {\textless\texttt{glyph$_i$}\textgreater}, where $T_i$ represents the $i$-th text character and {\textless\texttt{glyph$_i$}\textgreater} corresponds to the visual features of the glyph, as extracted in Section \ref{eff}.
Additionally, we append an initial empty JSON structure that retains only the essential keys.
This minimalist JSON setup facilitates the training process by providing a clear and straightforward framework for the model to learn from.

To construct the ground-truth response answers, we enhance the JSON structure by incorporating a comprehensive global layout description alongside a "detail" property.
The "detail" property specifies the detailed description for each individual glyph, leveraging the chain-of-thought methodology \cite{wei2022cot}.
This approach facilitates the management of complex and lengthy sequence layouts through a systematic, step-by-step reasoning process.
This enhancement not only boosts the accuracy of intricate layout generation but also ensures greater flexibility and responsiveness to a diverse range of user requirements.

\begin{table}[]
\centering
\resizebox{\columnwidth}{!}
{
    \begin{tabular}{ccccc}
    \toprule
    {\bf{Dataset}} & {\bf{Train}} & {\bf{Test}} & {\bf{Glyphs/img}} & {\bf{Total Glyphs}} \\
    \midrule
    TextLogo3K \cite{wang2021textlogo}  &  3170 & 300  & 4.69 & 16274              \\
    \rowcolor{gray!15}
    {\bf{SynTextLogo}}                  & 10000 & 1000 & 4.36 & 47963 \\
    \rowcolor{gray!15}
    {\bf{GenTextLogo}}                  &  6586 & 482  & 5.24 & 37036 \\
    \bottomrule
    \end{tabular}
}
\caption{Dataset statistics of existing dataset and our datasets. 
}
\label{dataset}
\end{table}

\begin{table*}[]
\centering
\resizebox{0.9\textwidth}{!}
{
    \begin{tabular}{l}
    \toprule
    {\bf{USER:}} \\
    \texttt{Can you arrange the given glyph images into a legible and aesthetically pleasing layout?} \\
    \texttt{Make sure there is no unnecessary overlap between the texts. Provide the results in JSON} \\
    \texttt{format, with each text's position and size represented by a bounding box in the format (left,} \\
    \texttt{top, right, bottom). Each coordinate should be a continuous number between 0 and 1.} \\
    \texttt{The glyph image for each text is "\text{T}\(_0\)": <glyph\(_0\)>, "\text{T}\(_1\)": <glyph\(_1\)>, \ldots} \\
    \texttt{The layout requirement is placing all elements in a horizontal line.} \\
    \texttt{Here is the initial JSON: [\textbraceleft "word": "\text{T}\(_0\)", "detail": "", "box": []\textbraceright, \ldots ] } \\
    \\
    {\bf{ASSISTANT: }} \\
    \texttt{There are N characters in this image, arranged horizontally from left to right ...} \\
    \texttt{[\textbraceleft "word": "\text{T}\(_0\)", "detail": "This character is the largest and positioned on the far left.",} \\
    \texttt{"box": [0.4759, 0.0231, 0.5863, 0.3790]\textbraceright, \ldots ] } \\
    \bottomrule
    \end{tabular}
}
\caption{
{\bf{Prompt templates}} for applying instruction fine-tuning on the content-aware text logo generation task. 
The placeholder $T_i$ and {\textless\texttt{glyph$_i$}\textgreater} will be replaced with character texts and glyph visual features during training and inference.
}
\label{layoutstructure}
\end{table*}

\section{Datasets}


Recognizing that existing text logo datasets may be insufficient for effectively fine-tuning Vision-Language Models (VLMs), we developed two comprehensive text logo datasets: {\bf SynTextLogo} and {\bf GenTextLogo}.
SynthTextLogo comprises logos generated from pre-defined templates crafted by skilled designers, ensuring a diverse range of styles and typographies.
In contrast, GenTextLogo features a curated collection of high-quality real-world logos from movie and commercial posters.
For each sample, we offer a comprehensive set of annotations, including bounding boxes, pixel-level masks, and character text recognitions, as well as detailed layout descriptions for every character.
The subsequent subsections outline our data processing pipeline, complemented by a detailed statistical analysis of the datasets in Table \ref{dataset}.
Examples are shown in Figure~\ref{intropdf}, with additional examples available in the supplementary materials.

\subsection{Synthesized Text Logo Layout}
We initially developed a synthesized dataset utilizing text rendering engines, which enabled us to generate diverse styled glyph images from a variety of TTF fonts.
Next, we collaborated with human designers to formulate several layout templates, arranging the rendered glyphs according to these templates to craft unique text logos using movie titles for character combinations.
To streamline the process, we implemented an automated pipeline that efficiently extracts binary masks and bounding box coordinates for each glyph.
While the complete collection comprises over 50k text logos, we selectively sampled 10k for training, ensures the balance between synthesized and general data.

\subsection{General Text Logo Layout}
To enhance data diversity, aesthetics, and align with real-world design scenarios, we have curated a comprehensive text logo dataset sourced from the internet.
This collection features text logos from movie posters and commercial product advertisements, showcasing titles that are intricately stylized and expertly designed, which boasts far more complex layouts compared to traditional text fonts and predefined templates.
To streamline the annotation process, we exclusively retain logos with RGBA channels, leveraging the alpha channel to effortlessly derive foreground masks.
For each glyph, we meticulously annotate binary masks and text characters, supplemented by the metadata of the logo image.
In total, we collect 7,068 logo samples, with 6,586 for training and 482 for testing.
In comparison to the previous dataset \cite{wang2021textlogo}, our collection offers greater diversity across logo types and styles, with longer text sequences that accommodate more complex design scenarios. 

\subsection{Textual Layout Description}
\label{texturallayoutdescription}
\begin{figure*}[h]
    \centering
    \includegraphics[width=0.8\textwidth]{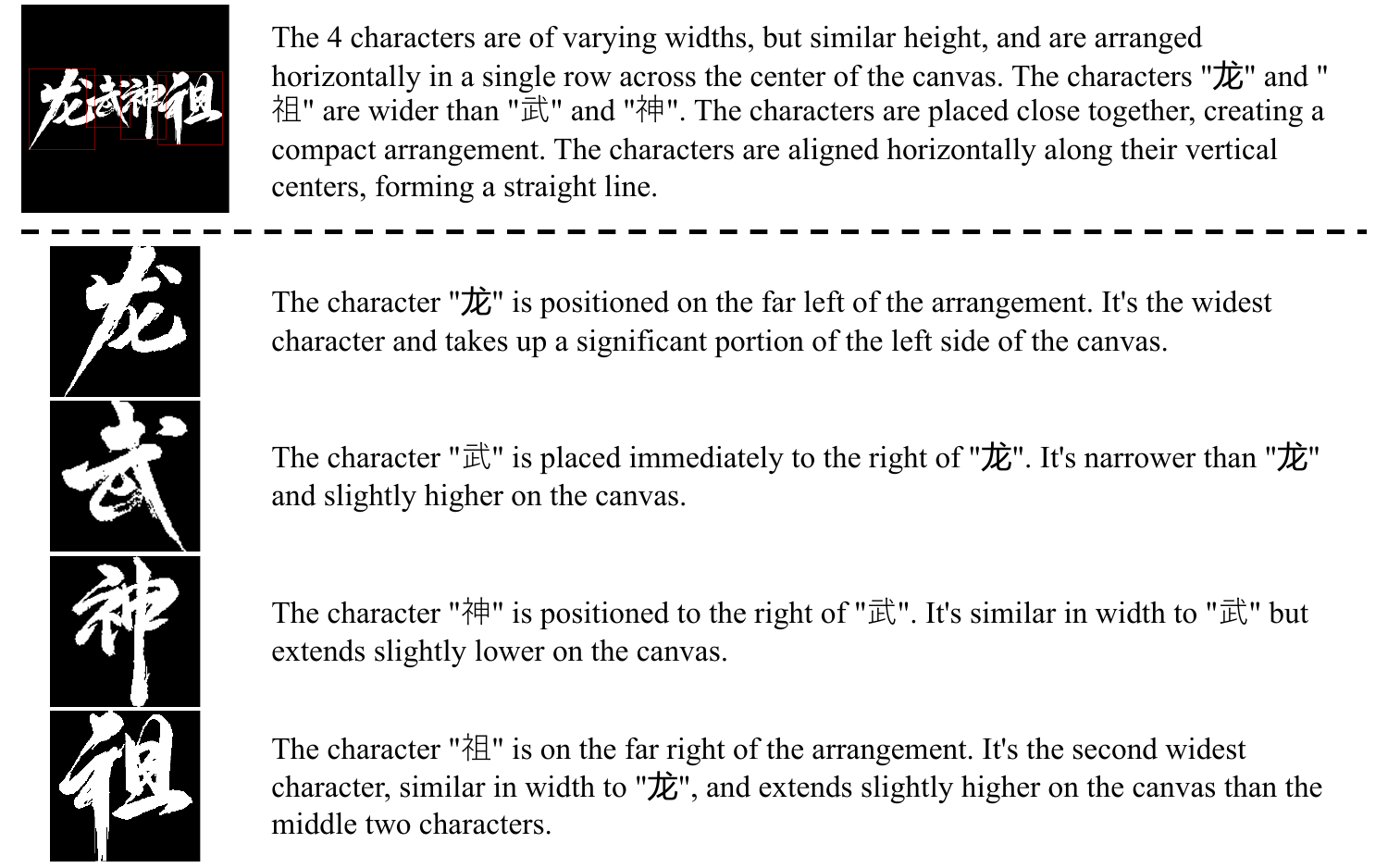}
    \caption{
    {\bf{Textual Layout Description}} complemented by our data processing pipeline in Section \ref{texturallayoutdescription}.
    We provide detailed textual descriptions for the overall text logo (top) and each text glyphs (bottom), with geometric relationships between glyphs.
    }
    \label{description}
\end{figure*}

In addition to providing spatial information for each glyph, we enrich our dataset with detailed textual descriptions, which enables a deeper understanding of the overall layout and the geometric relationships between glyphs, as illustrated in Figure \ref{description}.
Leveraging off-the-shelf VLMs \cite{claude, gpt4o, Qwen-VL, chen2023internvl}, we automatically generate detailed layout descriptions for each glyph within a text logo, alongside a comprehensive description of the logo's overall structure and style.
We then apply a rigorous filtering process to ensure the accuracy of these generated descriptions. 
This involves cross-verifying them with LLMs \cite{dubey2024llama3, gpt4o} by prompting with generated outputs and actual position coordinates, as well as validating through human annotators.
Only descriptions that accurately align with the glyphs are retained, with corrections made where necessary.
Most importantly, we ultimately extract and reconstruct user constraints for each logo based on the generated layout descriptions, paving the way for a user-friendly, constraint-conditioned layout generation scenario.

\section{Experiments}

\subsection{Implement Details}
\label{implementdetail}
Unless otherwise specified, we utilized the pre-trained 7B LLaVA-1.5 model \cite{liu2023llava}, with CLIP-L/336 serving as the glyph visual encoder.
Each input glyph image is resized to $336 \times 336$ pixels, and for these inputs, we implemented two distinct formats: square glyphs, which maintain the original aspect ratio by padding them to form a square, and tight glyphs, which are resized directly to fit the desired dimensions. 
During training, we randomly select the format type for each glyph input, which helps enhance the model's ability to perceive the intricate textural details of glyphs while preserving their original aspect ratio as closely as possible, thereby minimizing excessive distortion.
In contrast to the approach in \cite{wang2021textlogo}, which does not impose such constraints, our method seeks a balance between detail accuracy and aspect ratio fidelity.
This ensures that the generated glyphs remain visually coherent and faithful to their original design.
Once the layout results are generated, the complete logo is rendered using the differentiable composition techniques outlined in \cite{wang2021textlogo}, allowing for a seamless integration of the glyphs into the final layout.

To maintain training efficiency while preserving the model's generalization capabilities, we keep the vision encoder frozen and fine-tune only the LLM and the projection layers.
The newly introduced early feature projection layer is zero-initialized \cite{hu2021lora, zhang2023controlnet} to prevent the introduction of disruptive noise that could potentially compromise the stability of the pre-trained models.

We conduct training on 8 NVIDIA V100 GPUs, with a total batch size of 128.
The AdamW \cite{loshchilov2017adamw} optimizer with a learning rate of 2e-5 and a weight decay of 0.01 is employed, in conjunction with a cosine annealing scheduler that includes 3\% warm-up steps.
The training process integrates all three datasets for 7 epochs, which completes within 12 hours.

\begin{table}[tbp]
\resizebox{\columnwidth}{!}{
\begin{tabular}{lccccc}
\toprule
\multicolumn{1}{c}{{\bf{Methods}}} & FID $\downarrow$ & IS $\uparrow$ & IoU $\downarrow$ & V.B $\downarrow$ & Ratio $\downarrow$ \\
\midrule
Ground-Truth & - & - & 0.00 & 4.06 & 0.00 \\
rule (a) & 56.93 & 1.36 & 1.28 & 5.68 & 0.00 \\
rule (b) & 55.46 & 2.42 & 2.55 & 5.45 & 0.00 \\
TextLogo$\dagger$ \cite{wang2021textlogo} & 44.48 & 3.03 & 6.33 & 5.38 & 38.59 \\
\midrule
\rowcolor{gray!10}
{\bf{ours}} & 33.48 & 3.22 & 17.67 & 3.78 & 16.17 \\
\rowcolor{gray!10}
{\bf{ours}$\dagger$} & 38.73 & 3.14 & {\bf{6.21}} & 3.62 & 22.62 \\
\rowcolor{gray!10}
{\bf{ours (13B)}} & {\bf{33.77}} & {\bf{3.57}} & 14.12 & {\bf{3.52}} & {\bf{14.88}} \\
\rowcolor{gray!10}
{\bf{ours (13B)}$\dagger$} & 38.51 & 3.14 & 7.18 & 3.56 & 22.88 \\
\bottomrule
\end{tabular}
}
\caption{Quantitative results of {\bf{Unconstrained Layout Generation}} on the {\bf{TextLogo3K}} \cite{wang2021textlogo} dataset.
$\dagger$ indicates using tightly-cropped glyph images without keeping aspect ratios.
}
\label{textlogodataset}
\end{table}

\subsection{Evaluation Metric}
\label{evaluationmetric}
For a convenient comparison with prior approaches, we utilize the same evaluation metrics as \cite{wang2021textlogo} including Fr\'echet Inception Distance (FID) \cite{heusel2017fid} and Inception Score \cite{salimans2016inceptionscore}.
These metrics objectively assess the visual quality and diversity of the rendered logos from generated layouts.

However, the aforementioned metric falls short in fully capturing the geometric aesthetics of the generated logos, as it merely assesses the similarity between the generated and original outputs. 
To provide a more comprehensive evaluation of both geometric and aesthetic qualities, we additionally employ Overlap IoU, Visual Balance \cite{yang2024posterllava}, and Glyph Ratio Consistency as quantitative metrics.

The Overlap IoU metric evaluates the ratio of pixel-level overlap rather than bounding box-level intersections in document layouts \cite{hsu2023posterlayout, zhou2022cglgan}.
Visual Balance (V.B) \cite{yang2024posterllava} assesses the spatial equilibrium of the generated logo, which ensures that the logo's positioning is harmoniously centered within the canvas, contributing to an aesthetically pleasing design.
Glyph Ratio Consistency (Ratio) calculates the aspect ratio difference between the generated bounding box and the original glyph image, which is an essential metric to keep the original glyph's shape unchanged. 

Most importantly, to assess how well the model adheres to the specified input constraints, we sample a subset of 50 layouts from the test set and ask human annotators to verify if the generated layouts meet the given constraints. 
Following \cite{yang2024posterllava}, the average percentage of these violations is referred to the Constraint Violation Ratio (denoted as ViO).

\begin{table}[tbp]
\resizebox{\columnwidth}{!}{
\begin{tabular}{lccccc}
\toprule
\multicolumn{1}{c}{{\bf{Methods}}} & FID $\downarrow$ & IS $\uparrow$ & IoU $\downarrow$ & V.B $\downarrow$ & Ratio $\downarrow$ \\
\midrule
Ground-Truth & - & - & 0.00 & 3.92 & 0.00 \\
rule (a) & 60.23 & 1.01 & 10.47 & 5.21 & 0.00 \\
rule (b) & 56.44 & 1.38 & 12.98 & 5.94 & 0.00 \\
TextLogo$\dagger$ \cite{wang2021textlogo} & 57.82 & 1.06 & 13.55 & 5.47 & 42.68 \\
\midrule
\rowcolor{gray!10}
{\bf{ours}} & 26.22 & 2.38 & 19.84 & 3.78 & 16.72 \\
\rowcolor{gray!10}
{\bf{ours}$\dagger$} & 28.76 & 2.54 & 8.68 & 3.61 & 17.87 \\
\rowcolor{gray!10}
{\bf{ours (13B)}} & {\bf{25.23}} & 2.42 & 20.24 & 3.80 & {\bf{14.87}} \\
\rowcolor{gray!10}
{\bf{ours (13B)}$\dagger$} & 29.42 & {\bf{2.58}} & {\bf{7.86}} & {\bf{3.60}} & 17.52 \\
\bottomrule
\end{tabular}
}
\caption{Quantitative results of {\bf{Unconstrained Layout Generation}} on our {\bf{GenTextLogo}} dataset.
$\dagger$ indicates using tightly-cropped glyph images without keeping the aspect ratio.
}
\label{gentextlogodataset}
\end{table}

\begin{figure}[tbp]
    \centering
    \includegraphics[width=\columnwidth]{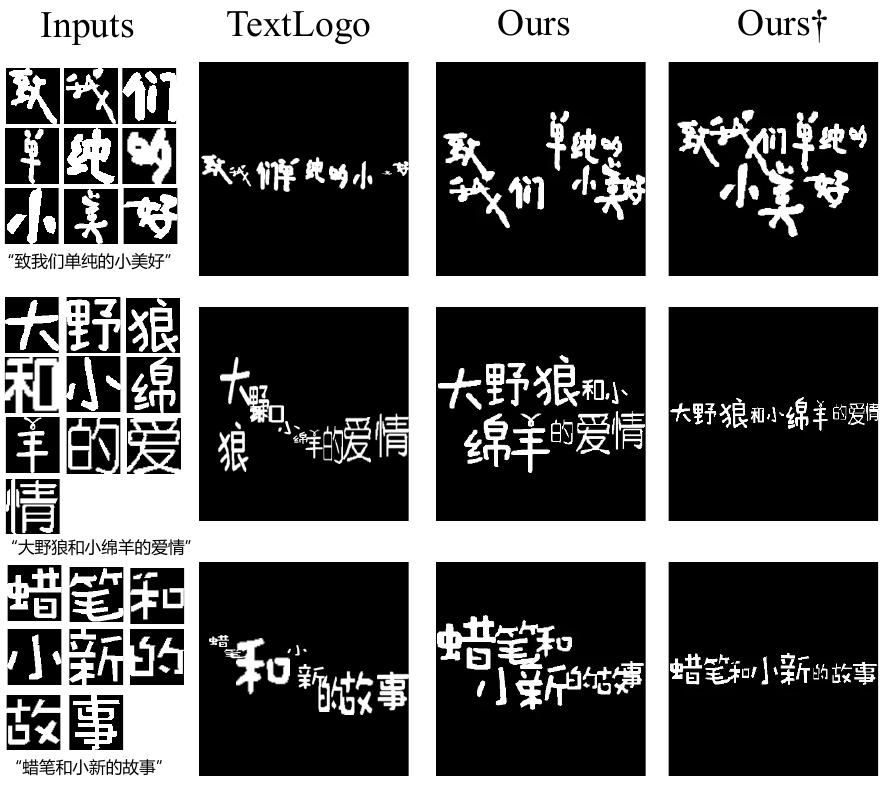}
    \caption{
    Qualitative comparisons of {\bf{Unconstrained Layout Generation}} on the {\bf{TextLogo3K}} \cite{wang2021textlogo} dataset.
    $\dagger$ indicates using tightly-cropped glyph images without keeping aspect ratios.
    }
    \label{textlogocompare}
\end{figure}

\subsection{Result Comparison}

\noindent{\bf{Unconstrained Text Logo Layout Generation.}}
We begin by assessing the capability to produce layouts without constraints across two distinct datasets, depicted in Table \ref{textlogodataset} and Table \ref{gentextlogodataset}.
Since our model is able to process two distinct formats of glyph images (as detailed in Section \ref{implementdetail}), for testing purposes, we utilize both square-padded and tightly-cropped glyph images, where the latter is marked with $\dagger$ for identification.
When evaluating, we remove the layout requirement part from the prompt template, as outlined in Table \ref{layoutstructure}, which allows the model full creative freedom in generating layouts.

In our experiments, we also incorporated two rule-based methods, labeled as rule (a) and rule (b) in Table 
 \ref{textlogodataset} and Table \ref{gentextlogodataset}, where glyphs are arranged (a) in a horizontal line and (b) in a horizontal or vertical line randomly.
The results illustrate that our model surpasses earlier methods, excelling in both logo quality and geometric aesthetics.

Impressively, our model consistently generates aesthetically pleasing layouts while preserving the original aspect ratio of glyphs, regardless of the input format. 
This capability offers a significant advantage in processing stylized glyphs, as illustrated in Figure \ref{textlogocompare}, underscores our model's perceptual ability to comprehend glyphs across various formats, thereby improving its generalization when handling unseen glyph shapes.

\begin{figure}[tbp]
    \centering
    \includegraphics[width=\columnwidth]{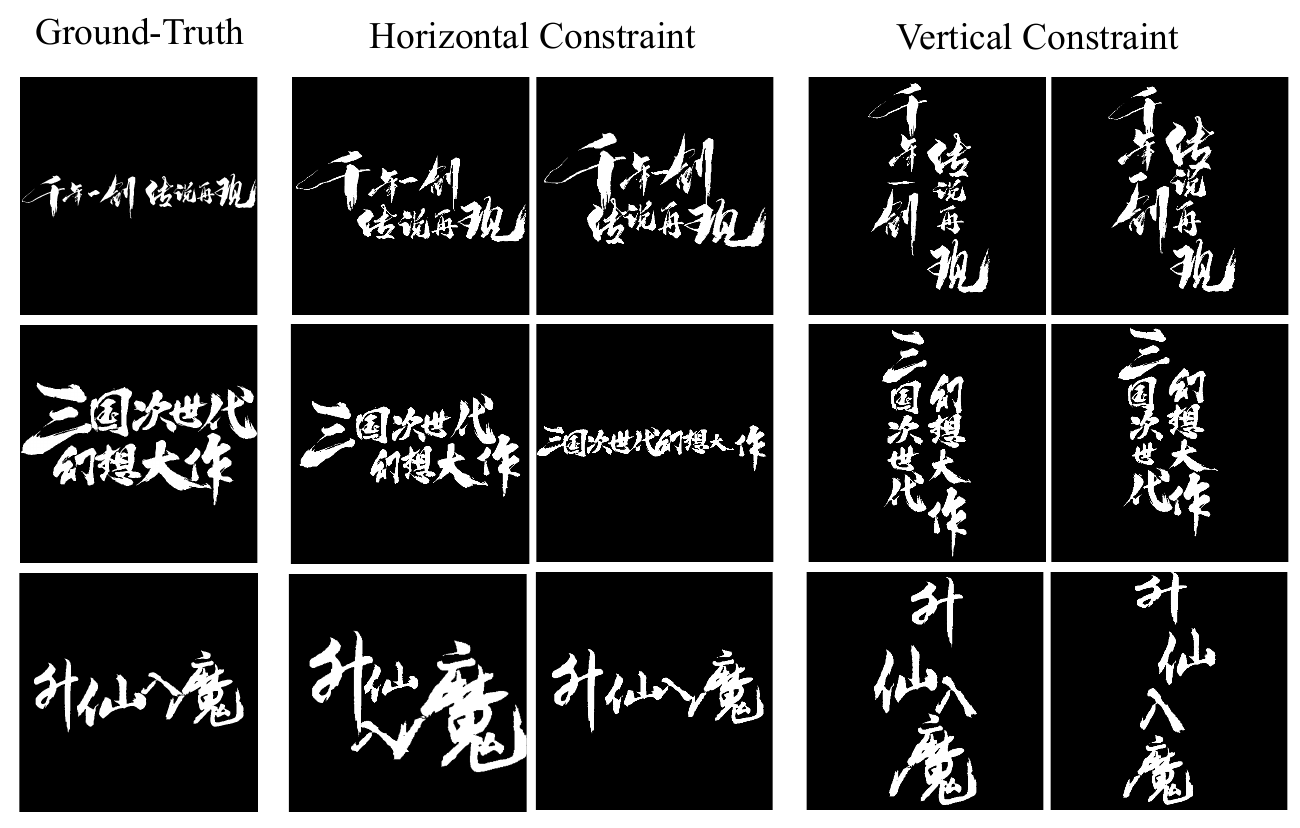}
    \caption{
    Qualitative results of {\bf{User-constrained Layout Generation}} on our {\bf{GenTextLogo}} dataset.
    }
    \label{constraint}
\end{figure}

\noindent{\bf{User-Constrained Text Logo Layout Generation.}}
We conducted additional experiments to assess the model's ability to generate outputs based on user-defined constraints, aiming to align the generation process more closely with real-world design scenarios.
Our findings are presented through both quantitative and qualitative analyses, as shown in Table \ref{userstudy} (utilizing the ViO metric) and Figure \ref{constraint}, respectively. 
Due to the limitations of \cite{wang2021textlogo}, which does not accommodate input constraints, we executed a prompt-tuning process to ensure a fair comparison.
The constraints are encoded using the pre-trained CLIP Text encoder \cite{radford2021clip}, and then integrated into the condition encoder described in \cite{wang2021textlogo}, serving as an additional condition to manage the constraints effectively.

However, as demonstrated by the ViO metric in Table \ref{userstudy}, the fine-tuned version of \cite{wang2021textlogo} still struggles to comprehend and adhere to user constraints, with a violation rate exceeding 50\%.
This indicates that the generated output layouts are essentially random, failing to respect the specified requirements.
In contrast, our model, which is enhanced with the LLM, exhibits a significantly superior ability to understand and implement user-specified constraints.
The improvements in layout generation are particularly evident in the visualization results shown in Figure \ref{constraint}, where our model successfully produces a wide variety of layout arrangements while rigorously adhering to the given constraints.
These results highlight not only the model's flexibility but also its precision in meeting user specifications.

\begin{figure}[tbp]
    \centering
    \includegraphics[width=\columnwidth]{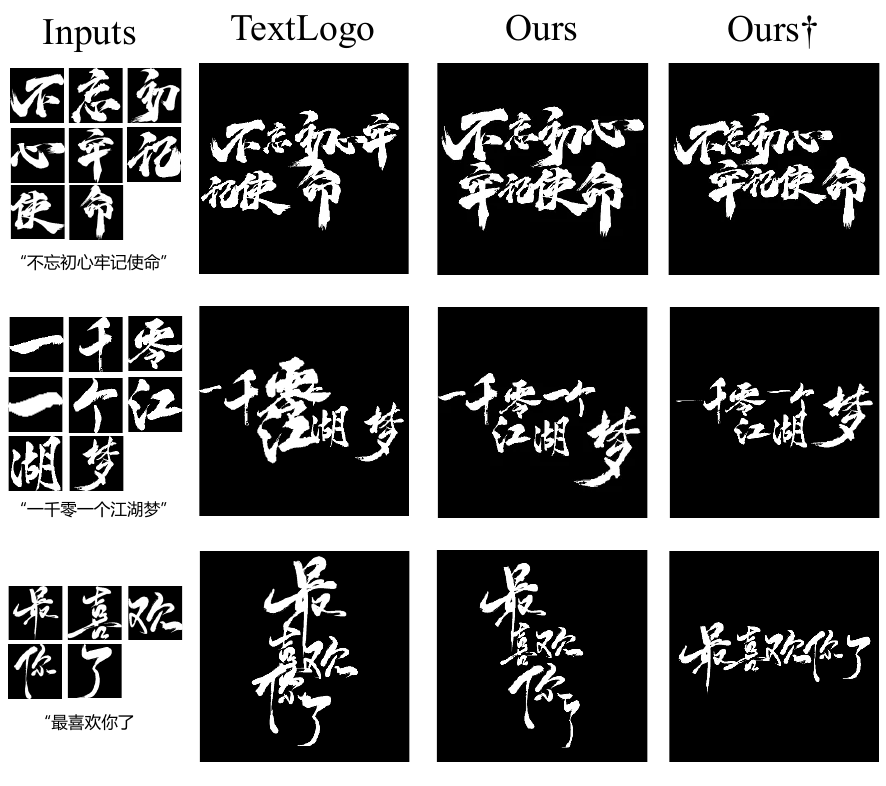}
    \caption{
    Qualitative comparisons of {\bf{Unconstrained Layout Generation}} on our {\bf{GenTextLogo}} dataset.
    $\dagger$ indicates using tightly-cropped glyph images without keeping aspect ratios.
    }
    \label{gentextlogocompare}
\end{figure}

\begin{table}[tbp]
\centering
\resizebox{0.8\columnwidth}{!}{
\begin{tabular}{lccc}
\toprule
Methods  & Prefer. $\uparrow$ & Quality $\uparrow$ & ViO $\downarrow$ \\ 
\midrule
TextLogo* \cite{wang2021textlogo} & 39.5\% & 2.5 & 0.57 \\
\rowcolor{gray!10}
{\bf{ours}}     & 68.3\% & 3.4 & 0.11 \\ 
\rowcolor{gray!10}
{\bf{ours (13B)}}     & {\bf{70.9\%}} & {\bf{3.9}} & {\bf{0.09}} \\ 
\bottomrule
\end{tabular}
}
\caption{User studies of {\bf{User-constrained Layout Generation}} on human annotators.
* denotes performing a prompt-tuning to accommodate input constraints.
}
\label{userstudy}
\end{table}

\subsection{Ablation Studies}
We conducted extensive ablation experiments to investigate the contributions of each model component, as summarized in Table~\ref{abaltion}.
By incorporating model adjustments such as Early Feature Fusion and Adaptive Average Pooling, we observed significant improvements, including a reduction in glyph collisions, as indicated by higher Intersection over Union (IoU) scores, and a decrease in computational requirements (+2.4\% with Early Feature Fusion and -86\% with Adaptive Average Pooling when processing an average of 1-10 glyphs).
Moreover, the inclusion of supplementary datasets and detailed textural layout description annotations led to substantial improvements in geometric accuracy, aesthetic quality, and layout diversity.

Additionally, augmenting the glyph inputs significantly improved aspect ratio consistency, which played a crucial role in preserving the original shapes of highly stylized glyphs.
This enhancement is vital for maintaining the integrity of complex designs, especially when dealing with intricate or artistic fonts.
By increasing the model's scale from 7B to 13B parameters, we observed further performance improvements, offering an optimal balance between computational efficiency and output quality.
pleasing results without compromising on speed or resource usage.

\begin{table}[tbp]
\resizebox{\columnwidth}{!}{
\begin{tabular}{lccccc}
\toprule
\multicolumn{1}{c}{{\bf{Methods}}} & FID $\downarrow$ & IS $\uparrow$ & IoU $\downarrow$ & V.B $\downarrow$ & Ratio $\downarrow$ \\
\midrule
baseline & 33.86 & 1.19 & 29.17 & 5.97 & 39.88 \\
\rowcolor{red!10}
+E.F.F & 30.46 & 1.21 & 27.66 & 4.63 & 34.91 \\
\rowcolor{red!10}
+A.A.P & 30.77 & 1.03 & 26.92 & 4.66 & 33.01 \\
\rowcolor{blue!10}
+SynTextLogo & 28.56 & 2.04 & {\bf{19.45}} & 3.88 & 20.81 \\
\rowcolor{blue!10}
+GenTextLogo & 26.22 & 2.38 & 19.84 & 3.90 & 16.72 \\
\rowcolor{blue!10}
+"detail" & 26.00 & 2.36 & 19.51 & {\bf{3.78}} & 16.94 \\
\rowcolor{red!10}
Scaling to 13B & {\bf{25.23}} & {\bf{2.42}} & 20.24 & 3.80 & {\bf{14.87}} \\
\bottomrule
\end{tabular}
}
\caption{Ablation studies of \textcolor{red!10}{\rule{6pt}{6pt}} model and \textcolor{blue!10}{\rule{6pt}{6pt}} training data.
E.F.F and A.A.P stand for Early Feature Fusion and Adaptive Average Pooling in Section \ref{modelarchitecture}.
"detail" stands for adding textural descriptions as illustrated in Table \ref{layoutstructure}.
}
\label{abaltion}
\end{table}

\subsection{User Studies}

In addition to evaluating the generative capabilities of our model through objective benchmarks, we conducted comprehensive user studies to provide a subjective comparison of the generated logo quality.
To ensure a robust evaluation of aesthetic preferences, we adopted a methodology similar to the ViO metric, as discussed in Section \ref{evaluationmetric}.
Participants were presented with pairs of logos and asked to select the one they considered to be of higher quality based on their personal preferences and visual appeal.
The sampled test data used in these studies included ground-truth logos sourced from the TextLogo3K \cite{wang2021textlogo} and GenTextLogo datasets, alongside our generated logos using identical text content but with different layout constraints applied to reflect various design scenarios. 

We present the preference rate, defined as the average percentage of instances in which participants considered the generated logos superior to the original ones.
Additionally, we report the average quality ratings (on a scale from 1 to 5, with 1 being the worst and 5 being the best) as evaluated by participants based on logo quality.
The results, summarized in Table~\ref{userstudy}, demonstrate that our model significantly outperforms previous methods when applied to real-world scenarios, reflecting its enhanced ability to produce visually appealing and user-preferred designs.

\begin{figure}[tbp]
    \centering
    \includegraphics[width=\columnwidth]{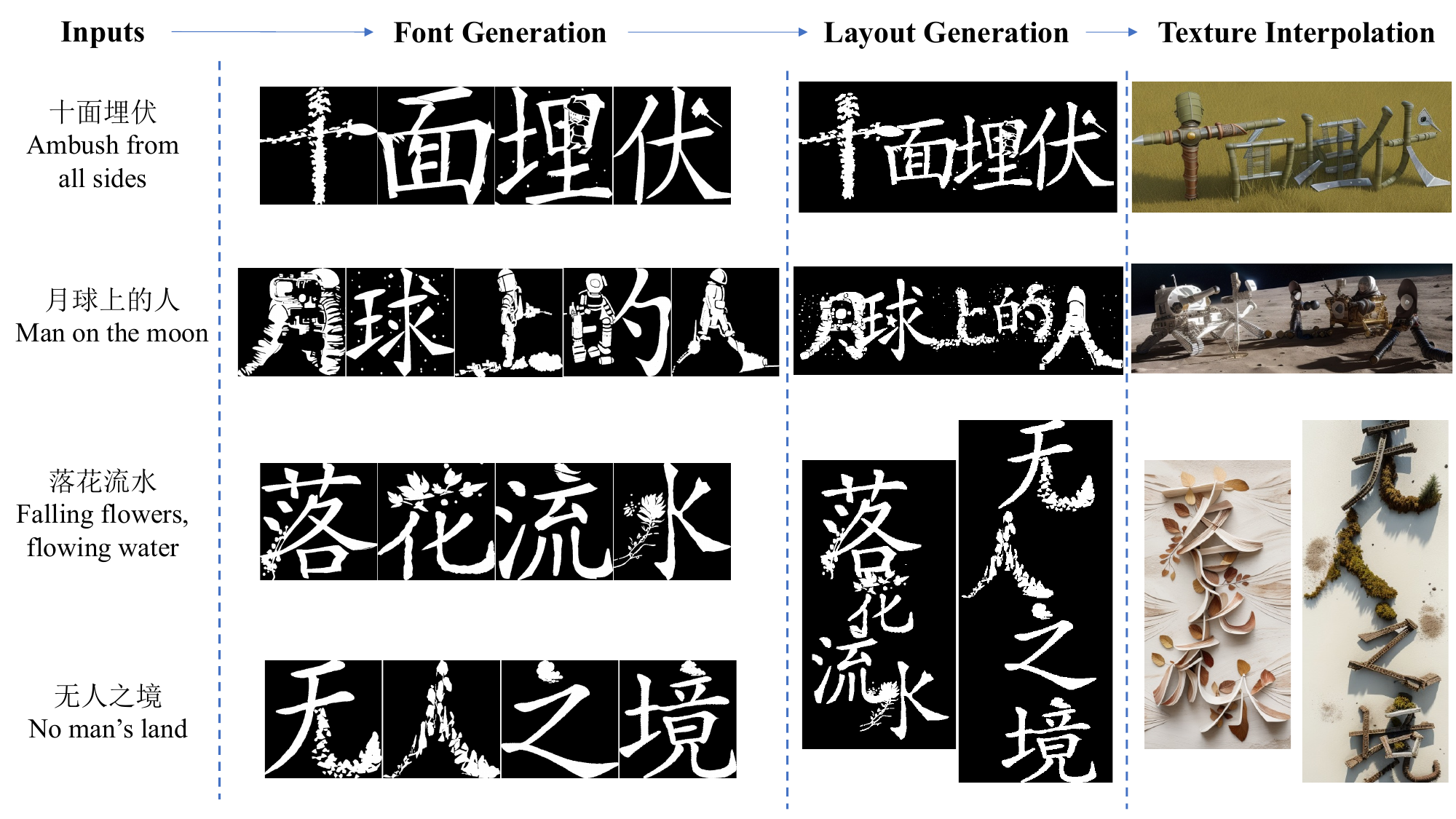}
    \caption{
    Pipeline of synthesizing text logos by combing with font generation and texture interpolation \cite{he2023wordart}.
    }
    \label{pipeline}
\end{figure}

We finally present a comprehensive pipeline for the automatic generation of text logos, integrating both font generation and textural interpolation \cite{he2023wordart}.
This process begins with the synthesis and rendering of font glyphs based on user-specified text inputs, followed by prompting our model to generate an optimal layout arrangement.
Subsequently, textural interpolation is applied to enrich the logos with semantic textures, elevating their visual appeal.
Examples of the complete pipeline are illustrated in Figure~\ref{pipeline}, with slight cropping applied for space efficiency.

\section{Conclusion}
We introduce {\bf GLDesigner}, a pioneering VLM-based designing framework content-aware text logo layout generation, capable of creating aesthetically appealing text logos by integrating multi-modal inputs with optional user constraints in natural language.
We propose two innovative techniques, {\bf Early Feature Fusion} and {\bf Adaptive Average Pooling}, which are designed to significantly reduce the computational load associated with processing multiple glyph images concurrently, leading to faster training and inference without compromising performance.
Additionally, we have constructed two comprehensive text logo datasets, {\bf{SynTextLogo}} and {\bf{GenTextLogo}}, which provide a rich source of training data essential for the instruction fine-tuning phase of our VLM.
Our model excels across various benchmarks and user studies, offering a stable and controllable generation of visually appealing text logos, thereby enhancing the overall user experience.

\bibliographystyle{ACM-Reference-Format}
\balance
\bibliography{sample-base}










\end{document}